%% file: paper.tex
\def\year{2021}\relax
\documentclass[letterpaper]{article} 
\usepackage{aaai21}  
\usepackage{times}  
\usepackage{helvet} 
\usepackage{courier}  
\usepackage[hyphens]{url}  
\usepackage{graphicx} 
\usepackage[switch]{lineno}
\usepackage{natbib}  
\usepackage{caption} 
\usepackage[utf8x]{inputenc}
\usepackage{verbatim}
\usepackage{stfloats}
\usepackage{amssymb,amstext,amsmath}
\usepackage{color}
\usepackage{booktabs}
\usepackage{algorithm}
\usepackage{multirow}
\usepackage{subfigure}
\usepackage{caption}
\usepackage[noend]{algpseudocode}
\usepackage{url}

\newcommand{\secref}[1]{Section \ref{#1}}
\newcommand{\figref}[1]{Figure \ref{#1}}

\newcommand{\tabref}[1]{Table \ref{#1}}

\newcommand{\cut}[1]{}
\newcommand{\tabincell}[2]{\begin{tabular}{@{}#1@{}}#2\end{tabular}}

\newfloat{figtab}{htb}{fgtb}
\makeatletter 
  \newcommand\figcaption{\def\@captype{figure}\caption} 
  \newcommand\tabcaption{\def\@captype{table}\caption} 
\makeatother
\title{Knowledge-Driven Distractor Generation for Cloze-Style Multiple Choice Questions}
\author{Siyu Ren, Kenny Q. Zhu\textsuperscript{\rm}\thanks{Corresponding Author.}}
\affiliations{
	
	Shanghai Jiao Tong University\\
	Shanghai, China\\
	roy0702@sjtu.edu.cn, kzhu@cs.sjtu.edu.cn
}
 
\begin{document}
\maketitle

\begin{abstract}
In this paper, we propose a novel configurable framework to automatically 
generate distractive choices for open-domain cloze-style multiple-choice questions. 
The framework incorporates a general-purpose knowledge base to 
effectively create a small distractor candidate set, and a feature-rich learning-to-rank model 
to select distractors that are both plausible and reliable. 
Experimental results on a new dataset across four domains show that 
our framework yields distractors outperforming previous methods both 
by automatic and human evaluation. 
The dataset can also be used as a benchmark for distractor generation research in the future.
\end{abstract}
\input{intro}
\input{method}
\input{experiment}
\input{related}
\input{conclusion}

\bibliography{paper}

\end{document}

%% file: intro.tex
\section{Introduction}
\label{sec:intro}
\begin{figure}[!htb]
	\centering
	\scalebox{1.0}{\includegraphics[width=1.0\columnwidth]{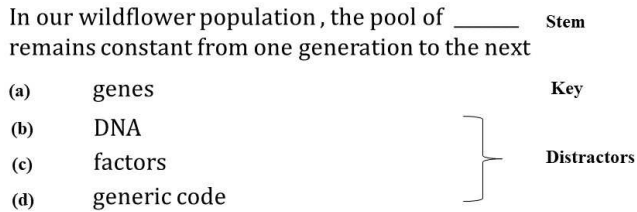}}
	\caption{A cloze-style MCQ} \label{fig:mcq}
	\end{figure}

Cloze-style multiple choice question (MCQ) is a common form of exercise used to evaluate the proficiency of language learners, frequently showing up
in homework and online testings.
Figure \ref{fig:mcq} shows a cloze-style MCQ, which typically 
consists of: a question stem with a blank to be filled in, 
the correct answer and multiple wrong answers used to distract testees.
Despite the high demand, manual crafting of such MCQs is highly time-consuming 
for educators, which calls for the automatic generation of 
as much practice material as possible from readily available plain texts so that formally usable quizzes can be generated after light-weight human calibration.

Distractor generation, which aims to generate distractive alternatives~(i.e., distractors) of the correct answer given the question stem, is a critical part of cloze-style MCQ construction. However, it is not only time-consuming but also non-trivial to produce appropriate distractors without rich experience in language education.

Literature in language pedagogy~\cite{haladyna2002review,pho2014multiple} generally recommends 
two criteria for designing distractors: 
{\em plausibility} and {\em reliability}. 
By plausibility, it means distractors should be semantically related to 
the key and grammatically consistent with the context given by stem 
to adequately discriminate learners' proficiency. 
By reliability, it means the distractor, when filled into the blank of
the stem, results in a logically incorrect or inconsistent statement.

Automatically generating distractors has been previously explored as part of cloze-style MCQ construction in a few studies. However, those methods generally assume prior knowledge of a specific domain~(e.g., science) of 
the given question and then use corresponding domain-specific vocabulary 
as candidate distractor set, ranked by various unsupervised 
similarity heuristics~\cite{sumita2005measuring,Kumar2015RevUP,jiang2017distractor,conceptEmb} or supervised machine learning model~\cite{sakaguchi2013discriminative,welbl2017crowdsourcing,liang2018distractor}.
Since identifying the concrete domain of each question and 
preparing large-scale domain-specific vocabulary require 
substantial human labor, such corpus-based methods cannot be easily 
applied in real-world scenarios.

Another issue is that previous approaches mainly focus on selecting plausible distractors 
while rarely adopt reliability checks to ensure that the generated 
distractors are logically incorrect. 
Despite some attempts in early approaches~\cite{sumita2005measuring,jiang2017distractor}, 
they both used it in the post-processing step to filter out candidate 
distractors rejected by diverse predefined filters such as syntactic feature~(e.g., role in the dependency parse tree), 
which may exclude useful distractors like \textit{DNA} in 
Figure \ref{fig:mcq}.


In this paper, we propose a configurable distractor generation framework for English cloze-style MCQ in the open domain, whose design is motivated by the shortcomings identified above. It mainly consists of two components: 
(1) a \textit{context-dependent} candidate set generator, which constructs a small set of 
candidate distractors from a general-purpose knowledge base, 
based on contextual information formed by the stem and the key; 
(2) a learning-to-rank model that takes both reliability checking and plausibility measures into consideration. By incorporating structured, human-curated
general-purpose knowledge base and conducting context-dependent 
conceptualization on the answer, we are able to effectively extract 
semantically-related candidate distractors without the need of 
domain-specific vocabulary. These candidate distractors are further re-ordered 
by a ranking model, trained with elaborately designed features to control the trade-off between plausibility and reliability.


Previous DG methods~\cite{Kumar2015RevUP,liang2017distractor,liang2018distractor} are evaluated either with sole human annotation or on ad hoc datasets that are often narrow in domain.
To the best of our knowledge, there is no open-source benchmark dataset for DG that is diverse enough to comprehensively evaluate the model performance. We compile a cross-domain cloze-style MCQ dataset covering science, trivia, vocabulary and common sense, which can be used as a benchmark for future research in DG. We further
investigate various instantiations of the framework. 

The contributions of this paper are three-folds:
\begin{itemize}
	\setlength{\itemsep}{1pt}
	\setlength{\parsep}{1pt}
	\setlength{\parskip}{1pt}
	\item we compile and open-source a diverse and comprehensive benchmark dataset for training and evaluating distractor generation model~(\secref{sec:data}).
	\item we propose a configurable distractor generation framework for open-domain cloze-style MCQ, which requires no domain-specific vocabulary and jointly evaluates the plausibility and reliability of distractors (\secref{sec:method}).
	\item we conduct comprehensive experiments to evaluate and analyze various instantiations of our framework and show that it consistently outperforms previous methods in both automatic ranking measures~(about 2\% F1 score) and human evaluation~(\secref{sec:endtoend}).
\end{itemize}

%% file: method.tex
\section{The Framework}
\label{sec:method}
\begin{figure}[th]
\centering
\scalebox{1.0}{\includegraphics[width=1.0\columnwidth]{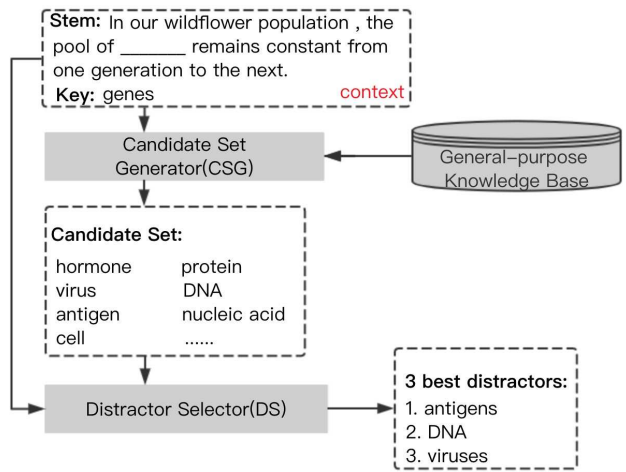}}
\caption{An overview of our framework.} \label{fig:framework}
\end{figure}

As illustrated in \figref{fig:framework}, our framework includes two 
components: \textbf{C}andidate \textbf{S}et \textbf{G}enerator~(\textbf{CSG}) and \textbf{D}istractor \textbf{S}elector~(\textbf{DS}). The first component CSG extracts candidate distractors that are semantically similar to the key from a general-purpose knowledge base~(KB). The second component DS, a generic feature-rich ranking model, then re-ranks those candidates according to more fine-grained assessment of grammatical consistency and reliability.

\subsection{Task Formulation}
Formally, given the stem $q$ and key $a$, the task of distractor generation is to generate $n$ most 
appropriate~(i.e., meeting the requirement of plausibility and reliability) distractors $d_i$ in a set of distractor-score pairs $D = \{(d_i, s_i)\}$, $1\leq i\leq n$,
in descending order of $s_i$.

\subsection{Candidate Set Generator (CSG)}
\label{sec:CSG}
The proposed CSG explicitly leverages the observation that distractors to 
an open-domain cloze-style MCQ are often words or short phrases 
living in a knowledge base (e.g., Probase~\cite{wu2012probase}, WordNet~\cite{leacock1998combining}) and stored as \textit{nodes} in a way that they are connected with the key through a common \textit{parent node} (which we refer to as \textit{concept} later). Instead of enumerating all words in a huge 
domain-specific vocabulary in early approaches, such hierarchical structure in knowledge base allows us to extract candidate distractors by only considering a reasonably small number of concepts $C$ that are semantically related to the key, 
which can be efficiently identified using KB-specific interfaces.

Nevertheless, the specific meaning of the key varies given different stems. For example, given the sentence: ``\textit{These survivors managed to swim to the \textbf{bank}},'' where \textit{\textbf{bank}} is the key, we would like to generate candidates like \textit{bay} rather than the more commonly used 
financial-related terms.

Inspired by the idea of context-dependent conceptualization~\cite{kim2013context}, we utilize a probabilistic topic model, LDA~\cite{Blei:2003:LDA:944919.944937}, to discover the latent topic distribution of the context as well as the topic distribution of all concepts in the concept set $C$\footnote{We have also experimented using cosine similarity from BERT-based embeddings but observed longer inference time and similar performance. More details are described in Appendix A.}. The posterior probability $p(c|a, q)$
of key $a$ belonging to concept $c$ conditioned on the stem $q$, is given by: 
\begin{align}
	p(c|a, q) &\propto p(c|a) \sum_{k=1}^K \pi_{a,q}^{(k)} ~ \gamma_{c}^{(k)}
	\label{eq:pc}
\end{align}
where $c$ is the concept, $\pi_{a,q}$ is the topic distribution of complete sentence formed by the stem and key, $\gamma_{c}$ denotes the topic distribution of concept $c$, $p(c|a)$ is the prior probability of $a$ belonging to $c$ corresponding to the specific choice of knowledge base, 
and $K$ is the total number of topics. 
Intuitively, concepts whose topic distribution resembles that of the complete sentence will be weighted higher than others.
\begin{figure}[t]
\centering
\scalebox{1.0}{\includegraphics[width=1.0\columnwidth]{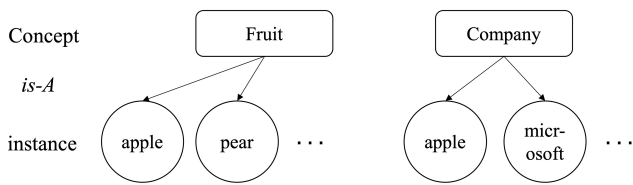}}
\caption{A snippet of Probase. Instances are connected to concepts under \textit{is-A} relation.} \label{fig:probase}
\end{figure}
After obtaining the conditional probability $p(c|a, q)$ of all concepts in 
$C$, by following the descending chain of \textit{is-A} relation and 
collecting hyponyms of these concepts as in \figref{fig:probase}, we get a probability distribution over 
all entities subsumed by the concepts in $C$:
\begin{align}	
	p_{i} = p(d_i|a,q) \propto \sum_{c\in C} p(d_i|c) p(c|a,q)
	\label{eq:pd}
\end{align}
where the probability $p(d|c)$ is also known as 
typicality~\cite{wu2012probase}. 
The prior probability $p(c|a)$ and typicality $p(d|c)$ can be used
off-the-shelf in some KBs (e.g., Probase) while for some other KBs
(e.g., WordNet) it is not the case, which endows our framework with 
the flexibility to be combined with a broad class of KBs and 
to be customized with different ways of calculating these two probabilities.

Then we remove candidates that occur in the stem and finally the top $m$ candidates with largest probabilities form a candidate distractor set $D_0 = \{(d_1, p_1), (d_2, p_2), \cdots,(d_m, p_m) \}$. 

\subsection{Distractor Selector (DS)}
\label{sec:DS}
Given the previously constructed candidate distractor set $D_0$, the final $n$-best distractors are generated in the following steps.
\subsubsection{Feature Extractor}
\label{sec:FE}
Given a triplet ($q$; $a$; $d$) where $q$ is the stem, $a$ is the key and $d$ is a candidate distractor, 
our DS first transforms it into a feature vector $f(q,a,d)\in \mathbb{R}^{33}$, in which the features are defined below:
\begin{itemize}
	\item[-] \textit{Embedding Similarity:} ~Similarity between $q$ and $d$ and similarity between $a$ and $d$ calculated using cosine similarity between their CBOW embeddings, which is effective for finding semantically similar distractors~\cite{guo2016questimator}. We use the average word embedding as the sentence embedding. 
\item[-] \textit{Contextual Embedding Similarity:} ~Cosine similarity between the ELMo~\cite{peters2018deep} embedding of $a$ and $d$. This feature is complementary to \textit{Embedding Similarity} since Word2Vec only capture static blended semantic of words, of which the significance is verified in our experiment.
\item[-] \textit{Morphological Similarity:} ~Edit distance, token/character length difference, singular/plural consistency, absolute and relative length of $a$ and $d$'s longest common prefix/suffix/subsequence. These features measure the morphological similarity and are useful for cases such as abbreviation 
(e.g., DNA and RNA). 
\item[-] \textit{POS Similarity:} ~Jaccard similarity between the 
POS tags of $a$ and that of $d$. The intuition is that a good distractor 
should share similar linguistic properties as the answer.
\item[-] \textit{Frequency:} ~Average unigram frequency of $a$ and $d$. Frequency has been previously utilized as a proxy for word's difficulty level~\cite{article}. This feature aids model to select distractors with similar difficulty as $a$.
\item[-] \textit{Compositional Similarity:} ~Jaccard similarity between token-level
unigram set and bigram set of $a$ and $d$. This feature is motivated by the observation that distractors might share tokens with answer.
	\item[-] \textit{Web-search Score:} ~Details of this feature are described later in this section.
\end{itemize} 

Features except \textit{Web-search Score} are integrated to mainly evaluate the plausibility of $d$ in various aspects and granularities. \textit{Web-search Score} is specifically introduced to assess
the validity of the sentence restored by each candidate in order to further strengthen reliability.  
First, search results are retrieved from the web by passing the full sentence 
concatenated from $q$ and $d$ to the Bing search engine automatically. 
Then, we use ReVerb ~\cite{fader2011identifying} to 
extract \textit{(argument1, relation phrase, argument2)} triplets 
involving $d$ from the sentence formed by $q$ and 
$d$, $\{t_{11}, t_{12}, \cdots, t_{1n}\}$, 
as well as triplets in the titles and snippets returned by the search engine, 
$\{t_{21}, t_{22}, \cdots, t_{2m}\}$. 
After that, we calculate embedding similarities 
between triplets and keep the maximal score, 
$T(q,d)$, that represents the correctness of triplet extracted from a sentence:
\begin{equation*}
	T(q,d) = max_{\tiny \begin{array}{c}
	i\in \{1,2,\cdots,n\} \\ j \in \{1,2,\cdots,m\}
	\end{array}
	} EmbSim(t_{1i}, t_{2j}),
\end{equation*}
where $EmbSim(t_{1i}, t_{2j})$ represents the Word2Vec-based cosine similarity between $t_{1i}$ and $t_{2j}$.
If $T(q,d)$ is small, then the sentence restored with the distractor $d$ is unlikely, thus $d$ should be a reliable distracter. 

\subsubsection{Ranker}
\label{sec:AMMR}
Given the feature vector $f(q, a, d)\in \mathbb{R}^{33}$ where $q$ and 
$a$ are the stem and key of triplet $(q; a; D_g)$ in the dataset, 
we propose to utilize a feature-based learning-to-rank model, 
which is trained in a supervised manner and learns to assign higher
scores to those $d$'s within the ground-truth distractor set $D_g$ 
than those in $D_0-D_g$. Reasonable distractors outside of $D_g$ are 
likely to be close to ground-truth distractors in the feature space 
$\mathbb{R}^{33}$, which can implicitly guide the ranker to learn relative 
ranking of negative examples during training.

Note that we do not restrict the ranker to be any specific model. 
One can choose to implement it using any state-of-the-art point-wise, 
pair-wise or list-wise learning-to-rank models. Theoretically, 
training a learning-to-rank model requires a relevance score associated with 
each distractor, which is not available in existing cloze-style 
MCQ dataset. We remedy this by setting the relevance score for $d\in D_g$ as 1 
and for $d\in \{D_0-D_g\}$ as 0. For point-wise ranker, 
it reduces into a binary-classifier~\cite{liang2018distractor}. 
The major difference between the point-wise ranking model and 
pair/list-wise ranking model is that the latter may learn latent 
feature patterns for discriminating between better or worse distractors 
through supervised training signal.


At test time, the ranking score $s_i$ for each candidate distractor $d_i$ 
predicted by the ranker is then used to sort the candidates in 
$D_0$ extracted by CSG and output the final $n$-best ranked list 
$D = \{(d_1, s_1), (d_2, s_2), \cdots, (d_n, s_n) \}$.

%% file: experiment.tex
\section{Experiments}
\label{sec:experiment}
In this section, we first present the dataset and evaluation metrics used in our experiments. Then we investigate several design choices of our framework 
and compare them against previous methods. Code at \url{https://github.com/DRSY/DGen}.

\subsection{The Dataset}
\label{sec:data}

\begin{figure}[ht!]
	 	\centering
 		\includegraphics[width=1.0\columnwidth]{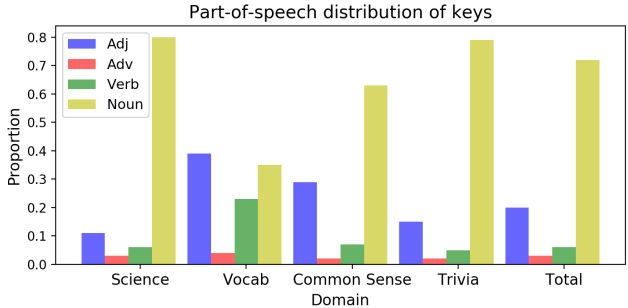}
 		\caption{POS distribution of keys.}
 		\label{fig:pos}
\end{figure}

Our MCQ dataset covers multiple domains including science,  
vocabulary, common sense and trivia.
It is compiled from a wide variety of open source  
MCQ datasets including SciQ~\cite{welbl2017crowdsourcing}, 
MCQL~\cite{liang2018distractor}, AI2 Science Questions as well as trivia, and 
vocabulary MCQs crawled from websites. 
We filter out MCQs whose keys are not single tokens since this paper only 
focuses on extractive cloze-style DG, resulting in 2,880 items in total among which 1176 are from SciQ, 300 are from MCQL, 275 are from AI2 and the rest from website resources.
Statistics of the dataset are summarized in \tabref{table:dataset} and \figref{fig:pos}.
We convert questions to cloze form by constructing Penn Treebank style trees using Stanford Parser~\cite{klein2003accurate}, replacing interrogative word with blank and adjusting node order according to the identified question type. The dataset is randomly divided into train/valid/test with a ratio of 8:1:1.
We use the TreebankWord tokenizer and POS tagger from NLTK~\cite{Loper:2002:NNL:1118108.1118117} to preprocess the stems and keys when constructing features.
\begin{table}[ht!]

	\centering
	\small
	\addtolength{\tabcolsep}{-2pt}
	\begin{tabular}{l|c|cccc}
	\toprule
	\multirow{2}{*}{Domain} & \multirow{2}{*}{Total} & \multirow{2}{*}{Science} & \multirow{2}{*}{Vocab.} & \multirow{2}{*}{\shortstack{Common\\Sense}} & \multirow{2}{*}{Trivia}\\
	& & & & & \\
	\midrule
	\# MCQs & 2880 & 758 & 956 & 706 & 460\\
	\midrule
	\# Distractors & 3.13 & 3.00 & 3.99 & 3.48 & 2.99\\
	\bottomrule
	\end{tabular}
\caption{Dataset Statistics (number of MCQs in each domain and average number of distractors per question).}
\label{table:dataset}
\end{table}
\subsection{Evaluation Metrics}
\label{sec:metrics}
\textbf{Automatic Evaluation.} ~Following~\cite{liang2018distractor}, we report F1 score(F1@3), 
precision(P@1, P@3) and recall~(R@3) to show how well the generated distractors 
match the ground truth distractors, 
as well as the mean reciprocal rank (MRR) and normalized discounted 
cumulative gain (NDCG@10). Sometimes the generated distractors
do not exactly match the ground truth, but are semantically very close. 
Word2Vec model trained on Wikipedia dump 
is utilized to measure the averaged cosine similarity~(Semantic Similarity@3) between the top three generated 
distractors and ground truth distractors.  

\noindent
\textbf{Human Evaluation.} ~Following~\cite{jiang2017distractor}, we ask three proficient English speakers to evaluate distractors' reliability and plausibility by showing them the key. We evenly sample 50 items in all domains from test set, each item contains multiple distractors including 3 generated by each method and all ground truth distractors 
designed by human experts. For each distractor, the judges decided whether it is correct or incorrect given the context. For a distractor deemed to 
be incorrect, the reliability score is 1 and the judges further assess 
its plausibility on a 3-point scale: 
``Obviously Wrong'' (0 point), 
``Somewhat Plausible'' (1 point), 
or ``Plausible'' (2 points). We then conduct an application-centric evaluation using another 50 samples without keys from test set by extending the original sample with additionally generated distractors and asking testees to answer it. The kappa inter-annotator agreement scores are 0.65 and 0.74 respectively for plausibility and reliability.



\subsection{Design Choices of CSG and DS}
\label{sec:ablation}

We investigate Probase and WordNet as the knowledge base in CSG and additionally extract all words and phrases from WordNet as a baseline of CSG in the following experiments. For Probase, both $p(c|a)$ and $p(d|c)$ are natively supported and can be obtained using official APIs. The size of a concept set $C$ is set to be 20. For nouns and verbs in WordNet, we treat the set of unique hypernyms~(as well as their siblings) of all synsets for $a$ as concept set $C$ and compute
$p(c|a)$ using the Laplace-smoothed Bayes rule on the lemma frequency provided in WordNet~(count on sense tagged text). We choose all synsets and their similar/antonymic sets as concept set $C$ for adjectives and adverbs in WordNet. Topic distributions $\pi_{a,q}$ and $\gamma_c$ are obtained using LDA pre-trained on Wikipedia dump and $K$ is set to 100.

For DS, we experiment with point/pair/list-wise ranking models to find the best practice.
Specifically, we employ AdaBoost~\cite{freund1997decision} as a point-wise ranker and LambdaMART~\cite{burges2010from} as both pair-wise and list-wise ranker. The dimensionality of feature vector $l$ is 33. Unigram frequency is calculated on Wikipedia dump. For the training of DS, negative examples are sampled using the top 100 candidates extracted by CSG excluding those that are within ground truths. At test time, DS takes as input the top 30 candidates extracted by CSG 
and 30 candidates sampled from WordNet's own vocabulary having the same POS tag. All hyperparameters are tuned on the dev set.

\subsection{Baselines}
We name our framework \textbf{CSG+DS} and compare it against the following baselines:
\begin{itemize}
	\item \textit{Thesaurus-based Method (TM)}~\cite{sumita2005measuring} ranks candidate distractors from synonyms of the key in WordNet-based on path similarity and applies post-filtering via IR.
\item \textit{RevUP}~\cite{Kumar2015RevUP} ranks candidate distractors based on weighted average of Word2Vec-based cosine similarity, dice coefficient and language model probability.
\item \textit{EmbSim+CF}~\cite{jiang2017distractor} combines Word2Vec-based cosine similarity, tri-gram and dependency candidate filtering in ranking and filtering respectively.
\item \textit{ED} use edit distance to measure the spelling similarity between distractors and key.

\item \textit{LR+RF}~\cite{liang2018distractor} combines logistic regression and random forest as a two-stage cascaded ranker with features measuring the plausibility of distractors.
\item \textit{LR+LM}~\cite{liang2018distractor} replaces random forest in LR+RF with LambdaMART.
\item \textit{BERT}~\cite{devlin2018bert} ranks candidates using cosine similarity of their BERT embeddings with that of the key.
\end{itemize}
Trigram and 5-gram Kneser Ney language model are built upon the original corpus of our dataset. 400-dimensional Word2Vec~(CBOW) is pretrained on Wikipedia dump and then fine-tuned on our corpus. Dependency tree is obtained using Spacy toolkit~\cite{spacy2}. We adopt the bert-base-uncased model and fine-tune it on our corpus with MLM objective.
\begin{table*}[t!]
	\small
	\centering
	\vspace{-0.4cm}
	\begin{tabular}{cc c c c c c c c}
		\toprule
		\multicolumn{2}{c}{\textbf{Instantiation}} &\multirow{2}{*}{F1@3} &\multirow{2}{*}{P@1} &\multirow{2}{*}{P@3} &\multirow{2}{*}{R@3} &\multirow{2}{*}{MRR} &\multirow{2}{*}{NDCG@10} &\multirow{2}{*}{\tabincell{c}{Semantic \\Similarity@3}} \\
		\\ [-1.8ex]
		\cline{1-2}
		\\ [-1.8ex]
		CSG &DS & & & & & & &\\
		\midrule
		\multirow{4}{*}{WordNet} &- &3.14 &3.49 &2.33 &5.43 &7.19 &8.66 &0.27 \\
		&point-wise ranker &7.26 &9.30 &5.55 &11.95 &14.30 &14.63 &0.36 \\
		&pair-wise ranker &7.11 &\textbf{10.07} &5.30 &12.14  &\textbf{14.40} &14.84 &0.35 \\
		&list-wise ranker &\textbf{7.71} &9.31 &\textbf{5.81} &\textbf{12.98} &14.34 &\textbf{14.94} &\textbf{0.36} \\
		\midrule
		\multirow{4}{*}{Probase} &- &5.88 &6.98 &4.39 &9.95  &12.07 &13.40 &0.35 \\
		&point-wise ranker &7.91 &8.14 &5.94 &12.98  &15.09 &17.69 &0.41 \\	
		&pair-wise ranker &\textbf{9.42} &10.08 &\textbf{7.00} &15.88  &17.33 &\textbf{19.70} &0.40 \\
		&list-wise ranker &9.19 &\textbf{10.85} &6.72 &\textbf{15.88}  &\textbf{17.51} &19.31 &\textbf{0.41} \\
		\midrule
		\multirow{4}{*}{w/o CSG} &- &- &- &- &- &- &- &- \\
		&point-wise ranker &5.59 &4.63 &3.98 &10.29 &8.67 &11.02 &0.36 \\
		&pair-wise ranker &5.62 &\textbf{5.01} &3.98 &10.10  &\textbf{9.28} &\textbf{11.60} &\textbf{0.36} \\
		&list-wise ranker &\textbf{5.94} &4.24 &\textbf{4.24} &\textbf{10.81} &8.81 &11.46 &0.35 \\
		\bottomrule
	\end{tabular}
	\caption{Comparison of combinations of different choices of CSG and DS. - means no ranking.}
	\label{table:instantiations}
\end{table*}
\begin{figure}[ht]
		\centering
		\includegraphics[width=1\columnwidth]{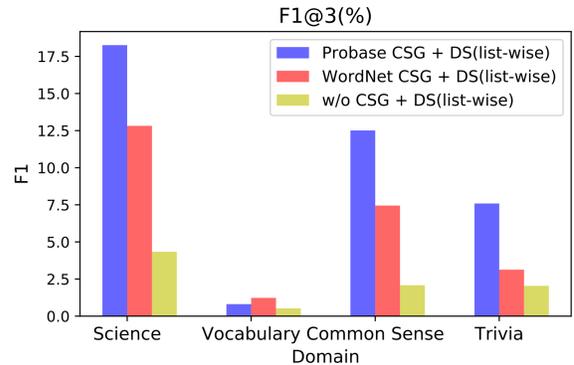}
		\caption{F1@3 score in different domains.}
		\label{fig:domains}
\end{figure}

%
\subsection{Results \& Analysis}
\label{sec:res}


\textbf{Combinations of CSG and DS.} ~\tabref{table:instantiations} shows the ranking performance for different combinations of CSG and DS. Without CSG, distractor selector trained with trivial negative examples is forced to select distractors from a rather large and noisy candidate set, therefore the performance is clearly worse. 
We also find that combining CSG with DS yields consistent improvement 
by all metrics and the improvement is more significant for WordNet CSG, 
which is mainly because $p(c|a)$ and $p(d|c)$ in WordNet are partly biased 
due to the limited scale of corpus they are estimated on, hence the 
supervised training will lead to more performance gain. Pair/list-wise ranker achieves comparable performance mainly due to the binarized relevance score.
Since named entities and common nouns mainly underpin Probase, 
DS with Probase CSG naturally get higher ranking scores
than its counterpart with WordNet CSG.




\noindent
\textbf{Domain Effect \& Feature Importance.} ~\figref{fig:domains} shows the F1@3 of CSG+DS in different domains. The performance drops most drastically when applied in the vocabulary domain because adjectives and adverbs in Probase and WordNet are either rare or not hierarchically organized. Another possible explanation is that the ground truth distractors in the vocabulary domain are less semantically-related to the key, which makes the learning process of the ranker oscillatory.
Our framework is especially better at generating distractors in science and commonsense domain, in which the keys and distractors are mostly subject-specific~(e.g. physics) terminologies, real-world entities and other common nouns. Trivia domain has similar characteristics but the keys are often rarer, therefore Probase suffers less due to its larger scope.
To have more insights on the proposed features, we also conduct a feature importance analysis of DS based on mean reduced impurity. It is defined as the total decrease in node impurity, weighted by the probability of reaching that node, averaged over all base classifiers. \tabref{table:feat_importance} reveals that semantic relation between $a$ and $d$ and web search score play a more critical role than features of other aspects.
\label{sec:endtoend}
\begin{table*}[t!]
	\footnotesize
	\centering
	\begin{tabular}{lc cc c c c c c cc}
		\toprule
		\multirow{3}{*}{\textbf{Method}} &\multicolumn{2}{c|}{\textbf{Human Evaluation}} &\multicolumn{7}{c}{\textbf{Automatic Evaluation}} \\
		\\ [-1.8ex]
		\cline{2-10}
		\\ [-1.8ex]
		& Reliability & Plausibility  &F1@3 & P@1 & P@3 &R@3 & MRR & NDCG@10 & \tabincell{c}{Semantic \\ Similarity@3}\\
		\midrule
		TM &95.57\%  &1.25$\pm$0.41 &1.74 &0.40  &1.16  &3.48   &2.69  &4.79  &0.21 \\
		\midrule
		\textbf{WordNet CSG} &98.66\% &1.25$\pm$0.34 &3.14 &3.49 &2.33 &5.43 &7.19 &8.66 &0.26 \\
		+~ED &90.66\% &1.26$\pm$0.41  &0.41 &0.12  &0.26  &0.58  &2.10  &1.93  &0.20 \\
		+~RevUP &93.65\%  &1.22$\pm$0.34  &4.07 &5.79  &3.21  &6.43 &9.31  &9.60  &0.32 \\
		+~EmbSim+CF &\textbf{99.12}\%  &1.21$\pm$0.49  &4.62  &6.17  &3.60  &7.40  &10.32  &10.94  &0.36 \\
		+~BERT &89.94\% &1.23$\pm$0.58  &5.68 &6.93  &4.23  &9.57  &11.10  &11.66  &0.30 \\
		+~LR+LM &96.66\%  &1.25$\pm$0.35 &6.48 &9.25  &4.89 &10.81   &13.42 &13.66 &0.29\\
		+~LR+RF &95.56\%  &1.25$\pm$0.38 &6.67 &8.10 &5.14 &10.81   &13.18 &13.73 &0.30 \\
		+~\textbf{DS(lise-wise)} &98.66\%  &\textbf{1.35$\pm$0.40}   &\textbf{7.71} &\textbf{9.31} &\textbf{5.81} &\textbf{12.98} &\textbf{14.34} &\textbf{14.94} &\textbf{0.36} \\
		\midrule
		\textbf{Probase CSG} &99.23\% &1.26$\pm$0.35 &5.88 &6.98 &4.39 &9.95  &12.07 &13.40 &0.34 \\
		+~ED &94.33\%  &1.23$\pm$0.38  &0.82 &1.16  &0.65  &1.30  &5.02  &4.92  &0.28 \\
		+~RevUP &94.87\%  &1.26$\pm$0.36  &6.27 &5.40  &4.63  &10.68  &11.74  &14.23  &0.37 \\
		+~EmbSim+CF &96.98\%  &1.19$\pm$0.47  &7.01 &8.10  &5.14  &12.34  &13.86  &16.33  &0.41 \\
		+~BERT &95.00\% &1.27$\pm$0.58  &7.05 &7.72  &5.14  &12.23  &13.60  &16.21  &0.36 \\
		+~LR+LM &98.98\%  &1.25$\pm$0.30   &7.62  &8.53  &5.81 &12.27 &15.56 &16.83 &0.40 \\
		+~LR+RF &99.13\%  &1.24$\pm$0.31   &7.48 &8.52 &5.42 &13.17   &15.87 &19.03 &0.40\\
		+~\textbf{DS(list-wise)} &\textbf{99.33}\%  &\textbf{1.30$\pm$0.34} &\textbf{9.19}  &\textbf{10.85} &\textbf{6.72} &\textbf{15.88} &\textbf{17.51} &\textbf{19.31} &\textbf{0.41} \\
		\midrule
		\textbf{w/o CSG} &- &- &- &- &- &-  &- &- &- \\
		+~ED &93.98\%  &1.00$\pm$0.12  &0.19 &0.38  &0.12  &0.38  &0.54  &0.53  &0.11 \\
		+~RevUP &92.88\%  &1.02$\pm$0.14  &2.01 &2.35  &1.35  &4.21  &3.95  &5.12  &0.38 \\
		+~EmbSim+CF &94.77\%  &0.93$\pm$0.52  &2.12 &2.70  &1.41  &4.24  &4.19  &5.24  &\textbf{0.42} \\
		+~BERT &93.87\% &1.02$\pm$0.24  &3.03 &2.88  &2.15  &5.14  &5.29  &6.78  &0.39 \\
		+~LR+LM &96.77\%  &1.05$\pm$0.28   &4.22  &4.34  &2.79 &8.69 &7.02 &10.16 &0.41 \\
		+~LR+RF &97.78\%  &1.02$\pm$0.20   &4.05  &4.21 &2.66 &8.55 &6.91 &10.08 &0.40\\
		+~\textbf{DS(pair-wise)} &\textbf{98.43}\%  &\textbf{1.06$\pm$0.14} &\textbf{5.59} &\textbf{5.01} &\textbf{3.98} &\textbf{10.10}  &\textbf{9.28} &\textbf{11.60} &0.36\\
		\midrule
		ground truth &\textbf{100\%}  &\textbf{1.41$\pm$0.35}  & - & - & - & - & - & - & -\\
		\bottomrule
	\end{tabular}
	\caption{End-to-end comparison on test set. - means no ranking algorithm to evaluate and ``ground truth'' denotes the score of ground-truth distractors associated with each item. Results are obtained by three runs with different random seeds.}
	\label{table:human}
\end{table*}
\begin{table}[t]
			\centering
		\footnotesize
		\addtolength{\tabcolsep}{-2pt}
		\begin{tabular}{c|c}
		\toprule
		Probase CSG & WordNet CSG \\

			\midrule
		contextual embed sim(a,d) & contextual embed sim(a,d) \\
		 word2vec embed sim(a,d)   & word2vec embed sim(a,d)  \\
		  word2vec embed sim(q,d)& word2vec embed sim(q,d)  \\
		  web search score & web search score \\
		  relative LCS len(d)   & relative LCS len(d)  \\
		  relative LCS len(a)  & relative LCS len(a)  \\
		  character len(d)   & character len(d)  \\
		  character len difference(a,d)   & character len difference(a,d)  \\
		  edit distance(a,d)   & edit distance(a,d)  \\
		  POS similarity(a,d)   &  relative common suffix len(a) \\
		\bottomrule
	\end{tabular}
	\caption{Top 10 important features of list-wise DS.}
	\label{table:feat_importance}
\end{table}
	\begin{table*}[t!]
		\footnotesize
		\centering
			\begin{tabular}{cccccccc} 
				\toprule
				 \textbf{Key} &RevUP &ED  & EmbSim+CF &BERT & LR+RF & LR+LM & \textbf{DS}\\
				\midrule
				  \color{red} \textbf{0.42} &0.06  &0.03  &0.04 &0.12  &0.11 &0.07   &\textbf{0.14}\\
				\midrule
			\end{tabular}
			\caption{Human evaluation on the frequency of being chosen as answers for each model paired with Probase CSG. DS denotes our list-wise distractor selector. Red colored number corresponds to the correct answer.} 
			\label{table:app}
		\end{table*}
\noindent

\textbf{End-to-End Comparison.} ~\tabref{table:human} shows the end-to-end results.
Despite the significantly reduced number of candidates, ranking methods with our candidate set generator can achieve much higher performance than with unstructured vocabulary. TM performs poorly due to its naive path similarity ranking criterion. The results of ED are worst among all unsupervised methods while embedding based methods can even achieve comparable performance against LR+LM/RF when provided with a high-quality candidate set.
BERT ranks distractors using contextualized representation thus leading to the lowest reliability according to human evaluation. 
LR+RF/LM achieves similar ranking performance yet obtain poorer reliability than CSG+DS since they only focus on the plausibility of selected distractors. CSG+DS, despite its relative simplicity, obtain consistent improvements over LR+RF/LM without two-stage cascaded training.
We observe certain inconsistencies between plausibility and automatic metrics of baselines, part of the reason may be that methods such as LR+RF/LM focus much on shallow feature patterns of ground-truth distractors and fail to unearth potential acceptable distractors.
However, distractors generated by CSG+DS yield highest-ranking measures while rated as most plausible by human annotators. 
Unsupervised methods work solely relying on the semantic similarity hence their reliabilities are generally lower than supervised ones, among which our DS turns out to be the most reliable. Exceptionally, EmbSim+CF gets higher reliability with WordNet, whose unreliable candidates get more chance to be eliminated by post-filtering than those in Probase.

\noindent
\textbf{Application-Centric Evaluation.} ~The frequency of generated distractors being chosen as answer for each tested model is shown in \tabref{table:app}. Our DS obtains the highest distracting rate compared to all baselines, indicating that distractors generated by our framework are more likely to distract testees in real-world scenarios. The Pearson correlation coefficient between the frequency and F1@3 is 0.46, implying a certain positive correlation between automatic metrics and actual distracting capacity.

\subsection{Case Study}
\begin{table*}[ht!]
	\footnotesize
	\centering
		\begin{tabular}{lcccccccc} 
			\toprule
			\# &Probase CSG  &RevUP &ED  & EmbSim+CF &BERT & LR+RF & LR+LM & \textbf{DS}\\
			\midrule
			1 &\color{red}protein   &\color{red}protein  &aldehydes  &starch &glycosaminoglycans  & hydrocarbon &methane    &\color{red} fat\\
			\midrule
			2 &alcohol  &alcohol  &carboxylic acid  &glycerol  &glycerol &methane  &\color{red}protein    &\color{red}protein \\
			\midrule
			3 &benzene  &amino acid  &alcohol   &\textbf{glucose}  &aldehydes &hormone  &hormone   &peptide \\
			\bottomrule
		\end{tabular}
		\caption{Top 3 distractors from different rankers running with Probase CSG
	given the stem ``The main source of energy for your body is \underline{\hbox to8mm{}}.'' and the key ``{\color{blue}carbohydrate}''. Red colored distractors are the ground truth, bold distractors are unreliable.}
		\label{table:example}
	\end{table*}
\tabref{table:example} compares predictions of all baselines and DS~(list-wise) running with Probase CSG. We can see that Probase CSG alone and RevUP are both able to generate distractors belonging to the same conceptual level as the key and accurately match one ground truth. However, running Probase CSG with ED yields 
distractors that are more semantically distant from the key. 
Despite the use of candidate filtering, EmbSim+CF still produces candidates 
like ``glucose'', which is an eligible answer to the stem. 
BERT instead generates compound names that are too technical and belong to a lower concept level than ground truth. 
Among all the supervised rankers, DS hits another ground-truth distractor 
``fat'' while LM+RF/LM predicts some obviously wrong distractors 
such as ``methane'' due to its coarse-grained features. A real-world mobile application using our framework can be found in Appendix B.


%% file: related.tex
\section{Related Work}
Extractive distractor generation typically involves two steps: 
candidate set generation and distractor selection.In the common scenarios, only the key and the stem are known beforehand and the set of candidates need 
to be automatically generated. A prior solution is to construct a distractor 
candidate sets from domain-specific vocabulary, 
thesauri~\cite{sumita2005measuring,smith2010gap} 
or taxonomies~\cite{mitkov2009semantic}. 
These domain-specific candidate sources are still not large or general enough to support open-domain distractor generation. In contrast, our framework is able to utilize a broad spectrum of general-purpose KBs and perform context-dependent conceptualization in an open-domain setting.

Previous approaches usually select distractors according to different metrics based on the key, including embedding-based similarities~\cite{guo2016questimator}, difficulty level~\cite{brown2005automatic,coniam2013preliminary}, WordNet-based metrics~\cite{mitkov2003computer} and 
syntactic features~\cite{agarwal2011automatic}.
Some approaches also consider the semantic relatedness of distractors with the whole stem ~\cite{pino2008selection,mostow2012generating} with domain restriction. Other researchers~\cite{liang2017distractor,liang2018distractor} investigate applying learning-based ranking models to select distractors that resemble those in actual exam MCQs, and quantitatively evaluate the top generated distractors. The DS in our framework incorporates a wide range of similarity measures to account for the plausibility in various aspects.

To generate reliable distractors, a supervised classifier~\cite{lee2007automatic} is adopted where they have a limited list of potential target words and distractors. 
Another way to perform reliability checking is by considering collocations 
involving the target word~\cite{smith2010gap,jiang2017distractor}. 
This approach is effective, but requires strong collocations statistics to 
discriminate between valid and invalid distractors and may not be applied to
the sentence in \figref{fig:mcq} which contains rare word combinations. 
A web search approach is applied by Sumita et al.~\shortcite{sumita2005measuring} to discard words that can be found on the web search results of the stem with blank filled by the distractor. We instead propose a novel web-based reliability checking feature and integrate it into DS for more accurate selection.

%% file: conclusion.tex
\section{Conclusion}
\label{sec:conclusion}
In this paper, we presented a configurable distractor generation framework for cloze-style open-domain MCQs. 
Using the proposed framework, we experimentally observe substantial performance gain in terms of distractor reliability and plausibility with less computational footprint. Depending on the characteristics~(e.g. capacity, POS distribution) of different general-purpose knowledge bases, the generated distractors may vary. Importantly, as knowledge bases with larger coverage and more advanced ranker inevitably emerge, they can be expediently integrated into our framework for further performance gain.

\section{Acknowledgement}
Sincere gratitude goes to Xinzhu (Elena) Cai for her initial
contribution and subsequent assistance in preparing this paper.